\documentclass[10pt,twocolumn,letterpaper]{article}

\usepackage{cvpr}
\usepackage{times}
\usepackage{epsfig}
\usepackage{graphicx}
\usepackage{amsmath}
\usepackage{amssymb}
\usepackage{mathtools}
\usepackage{algorithm}
\usepackage{algpseudocode}
\usepackage{pifont}

\usepackage{color}
\usepackage{amsfonts}
\usepackage{bm}
\usepackage{relsize}
\newcommand{\myMatrix}[1]{\bm{\mathit{#1}}}
\usepackage[utf8]{inputenc}
\usepackage[T1]{fontenc}

\usepackage[pagebackref=true,breaklinks=true,letterpaper=true,colorlinks,bookmarks=false]{hyperref}

\cvprfinalcopy 

\def\cvprPaperID{****} 

\ifcvprfinal\fi
\begin{document}

\title{Newton-type Methods for Inference in Higher-Order Markov Random Fields}

\author{Hariprasad Kannan \\
CentraleSup\'{e}lec-INRIA Saclay \\
Universit\'{e} Paris-Saclay\\
{\tt\small hkannan@gmail.com}
\and
Nikos Komodakis\\
Ecole des Ponts ParisTech\\
Universit\'{e} Paris Est\\
{\tt\small nikos.komodakis@enpc.fr}
\and
Nikos Paragios\\
CentraleSup\'{e}lec-INRIA Saclay\\
Universit\'{e} Paris-Saclay\\
{\tt\small nikos.paragios@ecp.fr}
}

\maketitle

\begin{abstract}
Linear programming relaxations are central to {\sc map} inference in discrete Markov Random Fields. The ability to properly solve the Lagrangian dual is a critical component of such methods. In this paper, we study the benefit of using Newton-type methods to solve the Lagrangian dual of a smooth version of the problem. We investigate their ability to achieve superior convergence behavior and to better handle the ill-conditioned nature of the formulation, as compared to first order methods. We show that it is indeed possible to efficiently apply a trust region Newton method for a broad range of {\sc map} inference problems. In this paper we propose a provably convergent and efficient framework that includes (i) excellent compromise between computational complexity and precision concerning the Hessian matrix construction, (ii) a damping strategy that aids efficient optimization, (iii) a truncation strategy coupled with a generic pre-conditioner for Conjugate Gradients, (iv) efficient sum-product computation for sparse clique potentials. Results for higher-order Markov Random Fields demonstrate the potential of this approach.
\end{abstract}

\section{Introduction}

Many computer vision problems can be modelled using Markov Random Fields ({\sc mrf}s). Maximum a posteriori ({\sc map}) estimation in an {\sc mrf} assigns labels to the nodes, in order to maximize their joint probability distribution. However, {\sc map} inference is NP-hard for a general graph and several approaches exist to achieve approximate solutions - \emph{graph cuts}, \emph{belief propagation} and \emph{LP relaxation} based methods \cite{surveyIJCV15}. In recent years, higher-order {\sc mrf}s have achieved excellent results in various applications, since they model far-reaching interactions between the nodes. While the topic of inference in pairwise {\sc mrf}s is well studied, development of scalable and efficient techniques for higher order models is evolving \cite{ishikawaPAMI11}, \cite{komodakisCVPR09}, \cite{kohliPAMI09}, \cite{fixCVPR14}, \cite{kolmogorovPAMI15}, \cite{vineetIJCV14}.

The LP relaxation approach has led to state-of-the-art algorithms and also, provides a theoretical foundation to the topic of {\sc map} inference. An attractive property of this approach is that it readily lends itself to inference in higher-order {\sc mrf}s \cite{komodakisCVPR09}, \cite{sontagUAI08}. Solving the LP relaxation in the primal is not scalable and the better approach is to solve the dual by exploiting the graph structure of the problem \cite{shlezingerCYBER76}, \cite{wainwrightTRANSIT05}. The various approaches to solve the dual can be categorized into coordinate optimization and gradient based methods. Block coordinate methods converge fast but can get stuck in sub-optimal corners when optimizing the non-smooth dual \cite{sontagUAI08}, \cite{kolmogorovPAMI15}. Since, we can recover better integer primal labels when closer to the optimum of the non-smooth dual, this can lead to poor solutions. On the other hand, supergradient methods can theoretically reach the global optimum \cite{komodakisPAMI11}. However, they have an $O(\frac{1}{\epsilon^2})$ rate of convergence towards an $\epsilon$-accurate solution.

These drawbacks can be addressed by approximating the dual with a smooth version. Smoothing the dual and applying accelerated gradient techniques was introduced by \cite{jojicICML10} and was further studied by \cite{savchynskyyCVPR11}, \cite{savchynskyyUAI12}. These algorithms reach an $\epsilon$-accurate solution in $O(\frac{1}{\epsilon}$) iterations. Block coordinate approaches can also work with a smooth objective \cite{meltzerUAI09}, \cite{hazanTransIT10}, \cite{meshiChap14}, which allows these algorithms to avoid sub-optimal corners. Some of these methods \cite{hazanTransIT10} can still get stuck, when the smoothing is reduced as we go closer to the optimum, in order to get more accurate results. Also, the scope for parallelization is more limited in block coordinate optimization methods compared to gradient based methods.

Alternating Direction Method of Multipliers (ADMM) inspired methods \cite{martinsJMLR15}, \cite{meshiMLKDD11} work with an augmented Lagrangian, which is also a smooth objective. However, convergence rate analysis for many of these methods has not been addressed and it is observed in practice, too. For example, AD3 \cite{martinsJMLR15} works well for many medium sized problems but can fail to converge in some instances. Also, scaling these methods to large vision problems with higher order cliques, has not been successful so far.

In recent years, Newton-type methods have led to state-of-the-art results in various Machine Learning problems \cite{schmidtPHD10}, \cite{martensICML10}, \cite{leeNIPS12}. These methods are able to take a better direction by considering curvature information and have quadratic convergence rate when sufficiently close to the optimum. One of the challenges while solving any optimization problem, is the conditioning of the objective. Intuitively, a problem is ill-conditioned if the change in objective value, due to a perturbation in variable value, varies radically depending on the direction of the perturbation. People have found that first order methods are faster when the condition number is small or moderate but second-order Newton-type methods perform much better for ill-conditioned problems \cite{fountoulakisTR15}. In {\sc map} inference, the smoothing has to be reduced as one gets closer to the opimum, leading to a considerably ill-conditioned dual objective.

The main disadvantage of Newton-type methods is the need to compute the Hessian, which can be very costly. However, it is worth investigating whether this is indeed the case or not for the problem in hand. Moreover, quasi-Newton methods which work with a Hessian approximation, can also lead to state-of-the-art results \cite{schmidtPHD10}, \cite{byrdSiamJO16}.

Our contributions are summarized as follows: (i) We show that for the smoothing based approach, it is possible to efficiently compute the Hessian and Hessian-vector product for a broad class of problems. (ii) We present a study of how to adapt Newton-type methods for {\sc map} inference in higher-order {\sc mrf}s. (iii) We demonstrate an efficient way to perform sum-product inference in chains of sparse pattern-based higher order cliques. This greatly enables the applicability of smoothing based approach for inference in higher order {\sc mrf}s. (iv) We showcase how Newton-type methods can beat first order methods for higher-order datasets. The remainder of this paper is organized as follows: section 2 outlines the concept of {\sc map} inference and the smoothing based approach. The main contributions of the paper are presented in sections 3 and 4, while experimental validation is in section 5, with a concluding discussion.

\section{MAP inference by optimizing a smooth dual} \label{Smooth dual main}
Consider a graph $\mathcal{G}$, where $V$ is the set of nodes, representing the random variables and $C$ is the set of \emph{cliques}, which enforce a certain relationship (e.g., smoothness) among the nodes they contain. Each node takes a label from a discrete set $l$. For example, object detection can be formulated using an {\sc mrf}, where each node corresponds to a part of the object, cliques represent geometric constraints between the parts and the labels are locations in the image. {\sc map} inference is the task of finding the labeling that maximizes the joint probability distribution of these random variables. It can be formulated as an equivalent energy minimization problem, as follows,
\begin{equation} \label{MRFEquation}
\text{minimize} \sum_{c \in C} \theta_c(\boldsymbol{x_c}) + \sum_{i=1}^n \theta_i(x_i)
\end{equation}
where $\theta_i(x_i)$ is the potential of node $i$ for label $x_i$ and $\theta_c(\boldsymbol{x_c})$ is the potential of clique $c$ for label $\boldsymbol{x_c}$. Cliques having more than two nodes become higher-order cliques.

Further, this energy minimization problem can be represented as an Integer Linear Program (ILP) as follows,
\begin{equation} \label{ILP}
\begin{gathered}
\text{minimize} \sum_{c} \sum_{\boldsymbol{x_c}} \theta_c(\boldsymbol{x_c})\phi_c(\boldsymbol{x_c}) + \sum_{i} \sum_{x_i} \theta_i(x_i)\phi_i(x_i) \\
\text{subject to} \sum_{x_{c \backslash i}} \phi_c(\boldsymbol{x_c}) = \phi_i(x_i), \;\;\; \forall c, i \in c, x_i \\
\sum_{x_i} \phi_i(x_i) = 1, \;\;\; \forall i; \quad \sum_{\boldsymbol{x_c}} \phi_c(\boldsymbol{x_c}) = 1, \;\;\; \forall c \\
\phi_i(x_i) \in \{0,1\}, \forall i, x_i; \;\;\; \phi_c(\boldsymbol{x_c}) \in \{0,1\} \forall c, \boldsymbol{x_c}.
\end{gathered}
\end{equation}
Here, $\phi_i(x_i)$ and $\phi_c(\boldsymbol{x_c})$ are indicator vectors for a given node $i$ (resp., clique $c$) for a label $x_i$ (resp., $\boldsymbol{x_c}$) and they are the discrete optimization variables. The constraints represent a polytope, called the \emph{Marginal} polytope. Relaxing the integrality constraint (last line), results in the LP relaxation, which is defined over the \emph{Local} polytope.

The Lagrangian dual of this LP relaxation is obtained through dual decomposition \cite{komodakisPAMI11}, \cite{sontagCHAP11}. The underlying idea is to decompose the graph into tractable sub-graphs, in order to derive the Lagrangian. This results in a concave, unconstrained and non-smooth optimization problem. Especially, if one treats each clique as the tractable sub-graph, then the approach given in \cite{sontagCHAP11}, leads to the following dual,
\begin{equation} \label{uncDD}
\begin{multlined}
\shoveleft{\underset{\delta}{\text{max.}} \sum_{c} \underset{\boldsymbol{x_c}}{\text{min.}} \big( \theta_c(\boldsymbol{x_c}) - \sum_{i:i \in c} \delta_{ci}(x_i)\big)} \\
\shoveright{ + \sum_{i}\underset{x_i}{\text{min.}} \big( \theta_i(x_i) + \sum_{c:i \in c} \delta_{ci}(x_i) \big)}
\end{multlined}
\end{equation}
Here the dual variables are $\delta_{ci}(x_i)$, for each label $x_i$ of each node $i$ within each clique $c$. If all the cliques have $k$ nodes each, with the nodes taking $l$ labels, then the total number of dual variables is $|C|.k.l$. We will denote this number as $\mathcal{N}$ and $\boldsymbol{\delta} \in \mathbb{R}^{\mathcal{N}}$, is the vector of dual variables. We note that the size of the dual is much lesser than the primal (\ref{ILP}), hence, there has been considerable research effort towards solving the {\sc map} problem through the dual.

In {\sc map} inference, there is an interplay between the sizes of the graph and the sub-graphs within dual decomposition \cite{komodakisCVPR09} \cite{wangICML13}. For medium sized graphs, decomposing into individual cliques, leads to excellent practical convergence. However, for large graphs, only bigger sub-graphs like chains of cliques lead to practical convergence \cite{komodakisCVPR09}, \cite{miksikBMVC14}. We would like to emphasize that the formulation shown in (\ref{uncDD}) decomposes the graph according to the cliques but in general, the sub-graphs can be bigger regions.

\subsection{Smooth dual} \label{Smooth dual}
The optimization of the non-smooth dual (\ref{uncDD}) has slow convergence and can stall at a point away from the optimum. In order to reach closer to the optimum of the non-smooth dual, optimization over smoother versions are carried out. The smooth dual can be obtained by adding to the primal objective (\ref{ILP}), entropies corresponding to all the nodes and cliques. The dual of this modified problem can be derived based on duality theory (refer \cite{meshiChap14}, \cite{jojicICML10} for further details). The smooth dual looks very similar to the non-smooth one, where each minimum operation in (\ref{uncDD}) is replaced by the negative soft-max function. The smooth optimization problem takes the following form,

\begin{equation} \label{smoothDD}
\begin{multlined}
\shoveleft{\underset{\delta}{\text{max.}} \sum_{c} \underset{\boldsymbol{x_c}}{\text{smin}} \big( \theta_c(\boldsymbol{x_c}) - \sum_{i:i \in c} \delta_{ci}(x_i); \tau \big)} \\
\shoveright{ + \sum_{i}\underset{x_i}{\text{smin}} \big( \theta_i(x_i) + \sum_{c:i \in c} \delta_{ci}(x_i); \tau \big)}
\end{multlined}
\end{equation}

We will denote the smooth dual function as $g(\boldsymbol{\delta})$. The negative soft-max function is defined as $\underset{x}{\text{smin}}(f(x); \tau) = \frac{-1}{\tau}\text{log}\sum_{x}\text{exp}(-\tau f(x))$. As $\tau$ increases, (\ref{smoothDD}) is a closer approximation of (\ref{uncDD}) and we get closer to the optimum of the non-smooth dual. However, a smooth approximation becomes increasingly ill-conditioned as it approaches the shape of the non-smooth problem (\S7, \cite{rockafellarBook09}). Hence, it is costlier to optimize for larger values of $\tau$. It is better to start with a very smooth version and switch to less smoothness, with warm start from the previous smoother version \cite{savchynskyyUAI12}.

\section{Trust-Region Newton for MAP Inference}

The central concept in Newton-type methods is the use of curvature information to compute the search direction. In each iteration, a local quadratic approximation is constructed and a step towards the minimum of this quadratic is taken. For the smooth dual $g(\boldsymbol{\delta})$, the equations to obtain the Newton direction are shown below. In these equations, $\myMatrix{B}(\boldsymbol{\delta}) \in \mathbb{R}^{\mathcal{N}X\mathcal{N}}$ carries the curvature information and can be either the exact Hessian ($\nabla^2 g(\boldsymbol{\delta})$) or a modified Hessian or a Hessian approximation (e.g., quasi-Newton).
\begin{alignat}{2} \label{newtonEqns}
& g(\boldsymbol{\delta} + \boldsymbol{p}) \approx g(\boldsymbol{\delta}) + \nabla g(\boldsymbol{\delta})^{T}\boldsymbol{p} + \boldsymbol{p}^{T}\myMatrix{B}(\boldsymbol{\delta})\boldsymbol{p} \\
& q(\boldsymbol{p}) = \nabla g(\boldsymbol{\delta})^{T}\boldsymbol{p} + \boldsymbol{p}^{T}\myMatrix{B}(\boldsymbol{\delta})\boldsymbol{p} \label{quadProb}\\
& \boldsymbol{p}^* = \underset{\boldsymbol{p}}{\text{argmin}} \ \nabla g(\boldsymbol{\delta})^{T}\boldsymbol{p} + \boldsymbol{p}^{T}\myMatrix{B}(\boldsymbol{\delta})\boldsymbol{p} \\
& \therefore \myMatrix{B}(\boldsymbol{\delta})\boldsymbol{p}^* = -\nabla g(\boldsymbol{\delta}) \label{linsys}
\end{alignat}
Here, the minimum $\boldsymbol{p}^*$ of the quadratic approximation $q(\boldsymbol{p})$ will be a descent direction for positive definite $\myMatrix{B}(\boldsymbol{\delta})$ and a line search along this direction determines the step size. For unconstrained problems, the Newton direction is found by solving the linear system (\ref{linsys}). Since, $\mathcal{N}$ is large for computer vision problems, we use Conjugate Gradients (CG) to obtain the Newton direction in our work. 

For a Newton-type method, if $\myMatrix{B}(\boldsymbol{\delta})$ is the exact Hessian ($\nabla^2 g(\boldsymbol{\delta})$) and if the backtracking line search tests full step length first, quadratic convergence is achieved, when sufficiently close to the optimum \cite{bertsekasBook99}. This is desirable, when compared to first order methods. As mentioned in section \ref{Smooth dual}, smoothing is reduced as the algorithm proceeds and this results in ill-conditioning. Newton-type methods have some robustness against ill-conditioning due their affine invariance, i.e., for some functions $f(\boldsymbol{x})$ and $\bar{f}(\bar{\boldsymbol{x}}) = f(\myMatrix{A}\boldsymbol{x})$, where $\myMatrix{A}$ is an invertible square matrix, the iterates of Newton-type methods will be related as $\bar{\boldsymbol{p}}_k = \myMatrix{A}\boldsymbol{p_k}$. Hence, in theory, these methods are not affected by ill-conditioning and in finite precision, the range of condition numbers in which Newton-type methods exibit robust behavior is more than that of first order methods. It is worth investigating how Newton-type methods handle {\sc map} inference.
 
\subsection{Hessian related computations} \label{hessComp}

Even though Newton-type methods have advantages, it may be expensive to populate and solve the linear system (\ref{linsys}). The computationally heavy steps for the Conjugate Gradient (CG) algorithm are: computing Hessian-vector products and regularly constructing and applying a preconditioning matrix. Generally, if the Hessian is difficult to populate, Complex Step Differentiation (CSD) \cite{csdLINK} can give very accurate Hessian-vector products, with each product costing roughly one gradient computation, which can be costly to be used within CG. So, we investigate the possibility of efficiently populating the Hessian and obtaining fast Hessian-vector products.

As mentioned in section \ref{Smooth dual main}, for medium sized problems it is sufficient to decompose according to cliques and achieve practical convergence. We will now show that for decompositions according to small sub-graphs like individual cliques, the Hessian can be computed very efficiently. In fact, it only takes time of about twice the gradient computation time to compute both gradient and Hessian together. 

\begin{figure}[t]
\begin{center}
  \includegraphics[width=0.8\linewidth]{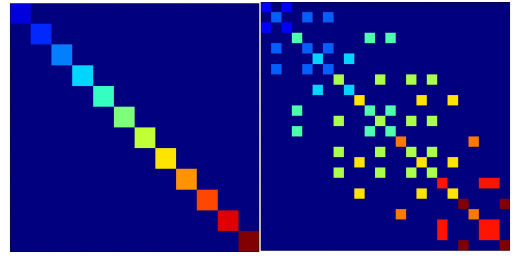}
\end{center}
   \caption{The two components of the Hessian. Within each component, blocks of the same color have the same values. In component one, there are as many unique blocks as cliques. In component two, each block-row/column has the same blocks.}
\label{hessianComponents}
\end{figure}

If we take a closer look at the Hessian, we observe that it can be written as the sum of two components, as shown in figure \ref{hessianComponents}. Both components have a block structure. Consider the following quantities,
\begin{alignat}{2} \label{hessianEqns}
& \mu_{ci}(x_i) = \frac{\underset{\boldsymbol{x_c}:\boldsymbol{x_c}(i) = x_i}{\sum} \text{exp}\big[ \tau.(\theta_c(\boldsymbol{x_c}) - \underset{n:n \in c}{\sum} \delta_{cn} (x_i))\big]}{\underset{\boldsymbol{x_c}}{\sum} \text{exp}\big[\tau.(\theta_c(\boldsymbol{x_c})-\underset{n:n \in c}{\sum}\delta_{cn} (x_n))\big]} \\
& \mu_{i}(x_i) = \frac{\text{exp}\big[ \tau.(\theta_i(x_i) + \underset{k:i \in k}{\sum} \delta_{ki} (x_i))\big]}{\underset{x_l}{\sum} \text{exp}\big[\tau.(\theta_i(x_l)+\underset{k:i \in k}{\sum}\delta_{ki} (x_l))\big]}
\end{alignat}

The elements of component one can be written as $H_{c,ij}(x_i,x_j) = \tau\big(\mu_{cij}(x_i,x_j) - \mu_{ci}(x_i)\mu_{cj}(x_j)\big)$, where $\mu_{cij}(x_i,x_j)$ can be calculated like $\mu_{ci}(x_i)$ by fixing the labels of two nodes. This leads to a block diagonal matrix, with as many unique symmetric blocks as there are cliques. The elements of component two can be written as $H_{st,i}(x_i,x_i) = \tau\mu_{i}(x_i)(1-\mu_{i}(x_i))$ and $H_{st,i}(x_i,x_j) = -\tau\mu_{i}(x_i)\mu_{i}(x_j)$. It has as many unique symmetric blocks as there are nodes and a given row or column has repeated copies of the same block. Here $c,s,t$ are cliques, $i,j$ member nodes, $x_i,x_j$ node labels and $\boldsymbol{x}_c$ clique labelling. Also, it is component two, which contains off-diagonal blocks, which is caused by overlapping cliques.

Hence, the elements of the Hessian can be computed by only iterating once through all cliques and nodes. Iterations corresponding to pairs of overlapping cliques is avoided. For the gradient, arrays of values need to be computed by iterating once through all cliques and nodes. For the Hessian, we need to compute (symmetric) blocks of values at each clique and node. Practically due to cache re-use, this is achieved overall with the overhead time of one gradient computation only. These blocks can be readily used for parallelizing Hessian-vector multiplication and we do it with simple OpenMP code. Thus, efficient Hessian-vector products leads to an efficient CG routine in our case.

For several problems, the Hessian is sparse because a clique overlaps only with a few other cliques. Nevertheless, because of the special structure, there is no need to exploit this sparsity for computing the unique blocks of the Hessian. Also, if there are many overlapping cliques, computing Hessian-vector products is not adversely affected because we will only have more copies of the same block along each row of the component two matrix, corresponding to the shared node. Hence, data movement in computer memory gets limited. Moreover, these data structures can be readily used for constructing preconditioners.

\subsection{Damping matrix approach}

The smooth dual (\ref{smoothDD}) is only strictly, not strongly, convex, i.e., away from the optimum, the Hessian is only positive semidefinite. In such a situation a trust-region approach is necessary to take meaningful Newton steps. Here, the same quadratic approximation (\ref{quadProb}) is minimized with the constraint $\lVert \boldsymbol{p} \rVert \leq \Delta$, where $\Delta$ is a trust radius. While developing our trust-region Newton method for {\sc mrf} inference ({\sc trn-mrf}), we addressed several issues: how to enforce the trust-region, how to improve speed of convergence by coping with the ill-conditioning and how to design a suitable preconditioner. It is a combination of these choices that lead to an usable algorithm and we note that, what works for {\sc map} inference, may not work for other tasks and vice-versa.

The Steihaug method seems like the first method to try for large problems. It has the nice property of shaping the trust-region into an ellipsoid, according to the landscape. This is achieved by minimizing (\ref{quadProb}), with the constraint $\lVert \boldsymbol{p} \rVert_{\myMatrix{M}} \leq \Delta$, where $\myMatrix{M}$ is a preconditioner and $\lVert \boldsymbol{p} \rVert_{\myMatrix{M}} = \boldsymbol{p}^{T}\myMatrix{M}\boldsymbol{p}$ \cite{connBook00}. However, in the absense of a good preconditioner, in a given outer Newton iteration, the algorithm quickly reaches the trust radius, before computing a good direction, leading to several outer iterations.

The Levenberg-Marquardt algorithm, which is generally used in least squares problems, offers another approach to impose a trust-region. The idea that we borrow is the damping matrix, which is a regularizer to address the strict convexity and the ill-conditioning. This matrix is added to the Hessian to obtain the modified Hessian in \ref{linsys}. The damping matrix restricts the Newton direction returned by CG to a trust-region. The Levenberg-Marquardt algorithm uses the scaled Hessian diagonal as the damping matrix. This choice gave very poor results for {\sc map} inference. Instead, we modify the Hessian as follows, $\myMatrix{B}(\boldsymbol{\delta}) = \nabla^2 g(\boldsymbol{\delta}) + \lambda \myMatrix{I}$, where $\lambda > 0$. While Steihaug works explicitly with a trust radius $\Delta$, we implicitly impose it through $\lambda$. This can be seen through these equations that tie the damping parameter $\lambda$ to a trust radius $\Delta$,
\begin{equation} \label{lambdaDelta}
\begin{gathered}
(\nabla^2 g(\boldsymbol{\delta}) + \lambda \myMatrix{I})\boldsymbol{p}^{*} = -\nabla g(\boldsymbol{\delta}) \\
\lambda(\Delta - \lVert \boldsymbol{p}^{*} \rVert) = 0
\end{gathered}
\end{equation}

In a sufficiently positive definite region (close to the optimum), $\lambda$ is close to zero and the Newton direction $\boldsymbol{p}^{*}$ is computed with the true Hessian. If $\lambda > 0$ then $\lVert \boldsymbol{p}^{*} \rVert = \Delta$, i.e., the direction is restricted by the trust-region. In ill-conditioned regions, $\lambda$ will be large and the enforced trust radius will be small. Thus, after every iteration, $\lambda$ is adapted and it can be done as follows,

\begin{equation} \label{lambdaAdjust}
\begin{gathered}
\rho < 0.25 : \lambda \leftarrow 2 \lambda; \;\; 0.25 < \rho < 0.5 : \lambda \leftarrow \lambda \\
0.5 < \rho < 0.9 : \lambda \leftarrow 0.5\lambda; \;\; 0.9 < \rho : \lambda \leftarrow 0.25\lambda  \\
\text{where,}\quad \rho = \frac{g(\boldsymbol{\delta} + \boldsymbol{p}) - g(\boldsymbol{\delta})}{q(\boldsymbol{p})-q(\boldsymbol{0})}
\end{gathered}
\end{equation}

$\rho$ signifies how well the quadratic of equation (\ref{quadProb}) approximates the dual ($g(\boldsymbol{\delta})$). As the algorithm approaches the optimum, $\lambda$ becomes vanishingly small and the algorithm reaches the true optimum without any perturbation.

\subsection{Forcing sequence for CG truncation}

Having found out the suitability of damping matrix based approach, it is still necessary to properly address the ill-conditioning caused by annealing. Trust-region Newton proceeds by taking approximate Newton steps, where in each outer iteration the run of CG iterations is truncated by a suitable criterion. However, as the algorithm approaches optimality, it is critical to solve the linear systems to greater accuracy and get better Newton steps \cite{schlickJCC87}. Otherwise, the algorithm will take too long to converge or will not converge at all. Generally, at an outer iteration $k$, CG can be truncated at iteration $j$ according to the following condition, $\lVert \boldsymbol{r}^j \rVert \leq \eta_{k} \lVert \nabla g(\boldsymbol{\delta}_k) \rVert $. Here, $\boldsymbol{r}^j = \myMatrix{B}(\boldsymbol{\delta}_k)\boldsymbol{\delta}_k^j + \nabla g(\boldsymbol{\delta}_k)$, is the residual of equation \ref{linsys}, at iteration $j$ of CG and $\eta_{k}$ is referred to as the forcing sequence. Through $\eta_{k}$ we reduce the residual and achieve more accurate Newton direction. For {\sc map} inference, we found textbook choices of the forcing sequence leading to Newton iterations not converging because the residual never becomes low enough to achieve the more accurate directions required for further progress. Hence, for {\sc trn-mrf}, the following criterion has been designed: $\eta_{k} = min(\frac{\epsilon_{\tau}}{k},\sqrt{\lVert \nabla g(\boldsymbol{\delta}_k) \rVert})$. This condition naturally imposes stronger condition on the residual for later iterations and also, the term ($\epsilon_{\tau}$), which takes smaller values as the annealing progresses, ensures that sufficiently accurate Newton steps are obtained, as smoothing reduces.

\subsection{Clique based Preconditioner and Backtracking search}

In order to further improve the efficiency of {\sc trn-mrf}, we will address one important aspect that influences trust-region Newton methods. This concerns the computational efficiency of the Conjugate Gradient routine. Convergence of CG iterations is a function of the number of distinct eigen value clusters of the linear system and a well preconditioned system ($\myMatrix{M}^{-1}\myMatrix{A}\boldsymbol{x} = \myMatrix{M}^{-1}\boldsymbol{b}$) will have fewer of clusters. Matrix $M$ should be as similar to $A$ as possible and should be efficient to construct and invert.

Standard preconditioning approaches like incomplete Cholesky, quasi-Newton and multigrid, fail to cope with generic {\sc map} inference problems. A clique based block diagonal preconditioner has along the diagonal, clique specific blocks ($\boldsymbol{H}^c$) of double derivative terms, i.e.,$\; H^c_{i,x_i,j,x_j} = \frac{\partial^{2} g(\boldsymbol{\delta}) }{\partial \delta_{ci}(x_i) \partial \delta_{cj}(x_j)} $, where $i, j$ are nodes belonging to the clique and $x_i, x_j$ are their labels. Since its structure is closely related to the {\sc map} inference problem, it performs quite well. The computational cost is of computing and inverting these blocks individually. This is done at the same time as computing the gradient and the Hessian. Applying this preconditioner corresponds to matrix-vector multiplications, involving the block inverses.

For sufficiently large values of $\lambda$, the modified Hessian is automatically well-conditioned and CG will converge fast. It is towards the optimum, when $\lambda$ reduces to vanishing values, that CG runs for large number of iterations. We observed considerable improvement in CG performance with this preconditioner and the results shown in this paper are based on this. Also, close to the optimum, there will be CG runs taking the maximum allowed number of iterations. We have set it as 250 for all our experiments.

The last aspect concerns backtracking search, once the Newton direction has been computed by the CG routine. When $\rho$ in equation (\ref{lambdaAdjust}) is less than a small value (say, $\epsilon_{\rho}$), it means a step of either very less decrease or an increase in the function value. So, we cannot directly take a step along this direction. However, it is desirable to take a step of sufficient decrease in every outer iteration of {\sc trn-mrf} and hence, we perform a backtracking search in these cases \cite{nocedalANL98}. The backtracking can be either done along a straight line or along a curved path (a subset of CG iterations). Although backtracking along a curved path gives good results in other problems ~\cite{martensICML10}, we observed poor results for {\sc map} inference: the final direction was very close to the steepest descent direction. On the other hand, performing a backtracking line search along the direction computed by CG, gave a huge speed-up for {\sc trn-mrf}. We have implemented a cubic interpolation based search, which is very efficient.

\subsection{Annealing schedule and stopping condition} \label{anneal_stop}

\cite{savchynskyyUAI12} suggests annealing $\tau$ by periodically computing the primal-dual (PD) gap of the smooth problem. For intermediate dual variables, they recover feasible primal variables by solving a small LP (called a transportation problem) for each clique. Since, their approach is adapted to pairwise graphs, they achieve results with less than thousand oracle calls (an oracle call is either an iteration or computation of the PD gap). However, for higher-order {\sc mrf}s, computing feasible primal variables and the primal objective is costly. Hence, computing PD gap of the smooth problem, every few iterations greatly affects computational efficiency.

We have used a simple but intuitive criterion for judging when to anneal. Since, we are working with a concave function, regions with lesser values of gradient euclidean norm are guaranteed to be closer to the optimum than regions with much greater values. Hence, if $\nabla g(\boldsymbol{\delta}_k)$ signifies the gradient after running $k$ iterations with a given $\tau$, we update $\tau \leftarrow \alpha \tau, 1 < \alpha $, if $\lVert \nabla g(\boldsymbol{\delta}_k) \rVert_2 < \gamma_{\tau}$. If we impose a strong enough threshold $\gamma_{\tau}$, we are guaranteed to achieve sufficient improvement for that particular $\tau$ and annealing can be performed. We have defined, $\gamma_{\tau} = \beta \lVert \nabla g(\boldsymbol{\delta}) \rVert$, just after $\tau$ is annealed. Similar to the proof in \cite{savchynskyyUAI12}, this annealing approach will work with any optimization algorithm that will converge to the global optimum for a fixed value of $\tau$. All the algorithms tested by us, have this guarantee for smooth, concave problems. Also, we ensure $\tau$ has reached a large enough value $\tau_{max}$ in order to obtain accurate results.

In order to exit the least smooth problem, \cite{savchynskyyUAI12} use the non-smooth PD gap. We have observed that {\sc trn-mrf}, due to its quadratic convergence rate, can exit based on classical gradient based condition itself. More precisely, with the $l_{\infty}$ norm and a threshold of $\zeta = 10^{-3}$ (\S 8, \cite{gillBOOK81}), {\sc trn-mrf} achieves good exit behaviour. However, first order methods take too long a time to achieve this gradient based exit condition and many times never do so. Hence, we have implemented a PD gap based approach, so that all algorithms can exit gracefully. The approach in \cite{savchynskyyUAI12}, is available only for pairwise graphs in the openGM library and implementing small LP solvers for all the higher order cliques looks challenging. \cite{meshiChap14} proposed a method involving only closed form calculations and we have implemented their approach.

Our complete trust-region Newton method, within an annealing framework, is described in Algorithm 1. In our experiments, we set, $\lambda_{0} = 1$, $\alpha = 2$, $\beta = \frac{1}{6}$, $\epsilon_{\rho} = 10^{-4}$, $\zeta = 10^{-3}$, $\tau_{max} = 2^{13}$ and $\epsilon_{\tau} = 0.1 \; \text{if} \; \tau < \frac{\tau_{max}}{4}, 0.01 \; \text{if} \; \frac{\tau_{max}}{4} < \tau < \frac{\tau_{max}}{2}, 0.001 \; \text{if} \; \frac{\tau_{max}}{2} < \tau$.

\begin{algorithm}
\caption{{\sc trn-mrf}: Trust-region Newton for {\sc map} inference}
\label{TRalgorithm}
\begin{algorithmic}[1]
\State Input: $\lambda_0, \tau, \tau_{max} > 0; \boldsymbol{\delta}_0 \in \mathbb{R}^{\mathcal{N}}; \alpha > 1.$
\State $\lambda \leftarrow \lambda_0, \gamma_{\tau} = \beta \lVert \nabla g(\boldsymbol{\delta}_0) \rVert_2$
\While {$\lVert \nabla g(\boldsymbol{\delta}_k) \rVert_{\infty} > \zeta \quad \text{or} \quad \tau < \tau_{max}$}
\If {$\lVert \nabla g(\boldsymbol{\delta}_k) \rVert_2 \leq \gamma_{\tau} \quad \text{and} \quad \tau < \tau_{max}$}
\State $\tau \leftarrow \alpha \tau; \;\;\; \gamma_{\tau} = \beta \lVert \nabla g(\boldsymbol{\delta}_k) \rVert_2 \;\;\; ;\;\;\; \text{adjust} \;\; \epsilon_{\tau}$
\EndIf
\State $ \myMatrix{B}(\boldsymbol{\delta}_k) = \nabla^2 g(\boldsymbol{\delta}_k) + \lambda \myMatrix{I}$
\State set $\eta_{k} = min(\frac{\epsilon_{\tau}}{k},\sqrt{\lVert \nabla g(\boldsymbol{\delta}_k) \rVert})$
\State Run CG while $\lVert \boldsymbol{r}^j \rVert > \eta_{k} \lVert \nabla g(\boldsymbol{\delta}_k) \rVert $
\State obtain Newton direction $\boldsymbol{p}$ and calculate $\rho$
\State update $\lambda$ according to equation (\ref{lambdaAdjust})
\State if $\rho < \epsilon_{\rho}$ then backtracking line search along $\boldsymbol{p}$ to obtain Newton step $\boldsymbol{p}_k$
\State $\boldsymbol{\delta}_{k+1} = \boldsymbol{\delta}_k + \boldsymbol{p}_k$
\EndWhile
\end{algorithmic}
\end{algorithm}

\section{Scalable Smoothing based approach}
Even though smoothing based approach has been scaled for large pairwise graphs \cite{savchynskyyUAI12}, higher order {\sc mrf}s with large label spaces and/or large graphs are less studied. In this section we show an efficient way to compute the smoothing operation with large label spaces for a very useful class of clique potentials. Next we demonstrate the use of quasi-Newton methods for problems with large graphs. 

\subsection{Pattern-based Smoothing} \label{sparseSP}
In the smooth dual (\ref{smoothDD}), we denote the first term as $g_{1}(\boldsymbol{\delta})$. It corresponds to cliques and involves log-sum-exp calculations over all possible labellings for each clique. This scales exponentially ($l^{k}$) with clique size $k$, where each node takes $l$ labels. Hence, computing the gradient of this dual is computationally heavy, especially for higher order cliques. \cite{jojicICML10} observed that these terms in the gradient correspond to marginal probabilities in a suitably defined graphical model. Hence, they can be computed using the sum-product algorithm \cite{kollerBook09}. Still, $O(l^{k})$ complexity remains.

Sparse, pattern-based clique potentials are very useful in computer vision \cite{komodakisCVPR09}, \cite{rotherCVPR09}. In these potentials, a big majority of the labellings take a constant (usually high) value and a small subset (of size $s$) take other significant values. Many clique potentials fit this description. \cite{komodakisCVPR09} showed an efficient way to perform max-product computation in chains of such cliques. It is not clear how to extend their work to sum-product. \cite{coughlanArxiv10} showed a sum-product approach in pairwise {\sc mrf}s. They achieved a complexity of $O(2l+s)$, instead of $O(l^{2})$. They mention that their idea can be used for higher order cliques with a factor graph representation but don't go into details. We have found that with a factor graph representation, their approach has a complexity of $O(l^{k-1}+l+s)$. Instead of a factor graph, if a clique tree representation (\S 10, \cite{kollerBook09}) is used, it is possible to achieve much better complexity of $O(l^{k-n}+l^{n}+s)$, where $n$ is the size of a subset of clique nodes. For individual cliques, these subsets will be (nearly) equal sized partitions of the clique. For clique chains, this is the separator set between two overlapping cliques. The separator set is the nodes shared by two cliques. For example, for cliques of size 4, with subset of size 2, one can quickly see the efficiency gain.

In the following, we will consider clique chains. This is for simple notation and these results are applicable to trees. Also, these results can be reduced to the case of individual cliques. While computing the gradient, the quantity $ \frac{\partial g_{1}}{\partial \delta_{ci}(x_i)}$ is equal to $p_{i}^{c}(x_i)$, i.e., the marginal probability of node $i$ taking label $x_i$ in clique $c$ of a suitably defined graphical model. Let $\psi_{c}(\boldsymbol{x}_c)$ and $\psi_{i}(x_{i})$ denote the clique and node potentials, respectively and $a$, $b$, $c$ be three neighbouring cliques in a chain, with $m$ and $n$ being the separator set between $a$ and $b$, $b$ and $c$, respectively. Sum-product message is passed from $b$ to $c$, as follows,

\begin{equation} \label{cliqPairMessage}
\begin{gathered}
m_{bc}(\boldsymbol{x}_{n}) = \underset{\boldsymbol{x}_{b \setminus n}} \sum \psi_{b}(\boldsymbol{x}_{b}) \nu(\boldsymbol{x}_{b \setminus n})\\
\text{where,} \;\; \nu(\boldsymbol{x}_{b \setminus n}) = \bigg( \underset{i \in b \setminus n} \prod \psi_{i}(x_{i}) \bigg) m_{ab}(\boldsymbol{x}_{b \setminus n})
\end{gathered}
\end{equation}

$m_{ab}(\boldsymbol{x}_{b \setminus n})$ is derived from the message sent by clique $a$ to $b$. This message was for the nodes in $m$ and $m_{ab}(\boldsymbol{x}_{b \setminus n})$ is obtained by marginalization (summation). Sometimes, $m = b \setminus n$. Let $\bar{\psi}$ denote the constant clique potential and let $Pat(\boldsymbol{x}_n)$ denote the pattern-based clique labellings corresponding to the seperator set labelling $\boldsymbol{x}_n$, then the first equation of (\ref{cliqPairMessage}) can be written as,

\begin{equation} \label{cliqPairPattern}
\begin{gathered}
m_{bc}(\boldsymbol{x}_{n}) = \underset{\substack{\boldsymbol{x}_{b \setminus n} \\ \in Pat(\boldsymbol{x}_n)}} \sum \psi_{b}(\boldsymbol{x}_{b}) \nu(\boldsymbol{x}_{b \setminus n}) + \underset{\substack{\boldsymbol{x}_{b \setminus n} \\ \notin Pat(\boldsymbol{x}_n)}} \sum \bar{\psi} \nu(\boldsymbol{x}_{b \setminus n})  \\
= \underset{\substack{\boldsymbol{x}_{b \setminus n} \\ \in Pat(\boldsymbol{x}_n)}} \sum \big( \psi_{b}(\boldsymbol{x}_{b}) - \bar{\psi}\big)\nu(\boldsymbol{x}_{b \setminus n}) + \bar{\psi} \underset{\substack{\boldsymbol{x}_{b \setminus n}}} \sum \nu(\boldsymbol{x}_{b \setminus n})  \\
\end{gathered}
\end{equation}

In (\ref{cliqPairPattern}), the second summation is computed only once ($O(l^{k-n})$) and used for all elements of $\boldsymbol{x}_{n}$. With a loop of complexity $O(s)$, the set of first summations for each element of $\boldsymbol{x}_n$ can be computed. Then iterating once through $\boldsymbol{x}_n$, gives the message in $O(l^n)$. Within this last loop to compute the message, the probabilities $p_{i}^{c}(x_i)$ of the corresponding nodes are computed. Also, the Hessian requires node pair marginals ($\mu_{c,ij}(x_i,x_j)$, section \ref{hessComp}) and they are calculated by parallelized sweeps of the sum-product algorithm working on independent subsets of the cliques.

\subsection{Quasi-Newton approach}

In section \ref{Smooth dual main}, we pointed out the trade-off between the sizes of the graph and the sub-graphs in dual decomposition. This becomes critical for large problems like stereo, where a decomposition based on individual cliques does not lead to practical convergence \cite{komodakisCVPR09}, \cite{miksikBMVC14}. Larger sub-graphs are needed and chains of higher order cliques is suitable for grid graphs. As mentioned in section \ref{sparseSP}, the Hessian for {\sc map} inference requires node pair marginals and it is very costly to calculate node pair marginals for clique chains. Hence, it is not practical to populate the Hessian for problems which require large sub-graphs. Since we have promising results with trust-region Newton for medium sized problems with individual clique based decomposition, we explore quasi-Newton methods for large problems. Note that quasi-Newton methods have only super-linear convergence rate when sufficiently close to the optimum.

Quasi-Newton methods build an approximate Hessian by low rank updates of an initial approximation (usually, a scaled Identity matrix) with vector pairs, $\boldsymbol{s}_k \triangleq \boldsymbol{\delta}_{k+1} - \boldsymbol{\delta}_{k}$ and $\boldsymbol{y}_k \triangleq \nabla g(\boldsymbol{\delta}_{k+1}) - \nabla g(\boldsymbol{\delta}_k)$. For large problems a limited memory variant is required, based on the most recent $m$ pairs of vectors. One can either approximate the Hessian or the Hessian inverse. For not strongly convex and/or ill-conditioned problem, the Hessian inverse approximation leads to large Newton steps. Downscaling them to a trust region radius does not help, as the direction itself is poor to start with. A trust-region based approach using the Hessian approximation is the better choice. Hence, we take the same approach as Algorithm 1 with an Hessian approximation (based on L-BFGS) in the place of $\nabla^2 g(\boldsymbol{\delta}_k)$.

The Hessian approximation is of the form of an Identity + rank-$r$ matrix. The Conjugate Gradient algorithm converges in $O(r)$ iterations for such a linear sytem (\S 11.3.4, \cite{golubBook12}). Morever, matrix-vector products in quasi-Newton can be obtained through a compact representation \cite{byrdMP94}. Thus the cost for the CG routine is a very small fraction of the cost of computing the gradient ($0.5\%$ in our experiments). Thus, given the iteration cost of quasi-Newton being comparable to first order methods, it is worthwhile to check whether quasi-Newton converges faster to the optimum.

\section{Experiments} \label{Experiments}
We present results based on higher-order {\sc mrf} models. As our baseline, we used {\sc fista} with backtracking line search \cite{beckJIS09}, Smooth Coordinate Descent ({\sc scd}) based on Star update \cite{meshiChap14} and {\sc ad3} \cite{martinsJMLR15}. Note that Star update based SCD, consistently performed better than Max-Sum Diffusion based SCD in our experiments.

First, we tested on medium sized problems and compared {\sc trn-mrf}, {\sc fista}, {\sc scd} and {\sc ad3}. We formulate the problem of matching two sets of interest points as {\sc map} inference, based on \cite{duchennePAMI11}. For $n$ points, each point in the source can be matched to any of the points in the target, i.e., $n$ labels. The {\sc mrf} is constructed by generating $4n$ triangles in the source and each triangle in the source and target are characterized by tuples of side length. For each source tuple, we find the top 30K nearest neighbours among the target tuples. 30K is much lesser than all possible triangles in the target and this is a sparse, pattern-based clique potential. The higher-order cliques potentials are defined as $\exp(\frac{-1}{\gamma}((s_a-t_a)^{2}+(s_b-t_b)^{2}+(s_c-t_c)^{2}))$, where $s$ and $t$ refer to source and target, respectively and $\gamma$ is the average squared differences between source tuple and its 30K nearest neighbours in the target. To get state-of-the-art results additional terms to disallow many-to-one mapping will be needed. For the sake of simplicity we ignore this issue.

We first tested with a synthetic problem, with $n=81$ on the 2D plane. We added Gaussian noise to these points to create the target image. The unary potentials are defined by assigning the value $i$ to node $i$ in both the images and taking the absolute differences. The results are presented in table \ref{synthTable}, where two levels of added noise were tested.

\begin{table*}[t]
\resizebox{\textwidth}{!}{
  \centering
  \begin{tabular}{ | l | l | l | l | l | l | l | l | l | }
    \hline
     & \multicolumn{4}{c|}{$\sigma = 0.5$} & \multicolumn{4}{c|}{$\sigma = 0.8$} \\ \hline
    Algorithm & TRN & SCD & FISTA & AD3 & TRN & SCD & FISTA & AD3 \\ \hline
    time (seconds) & 948.9 & 1742.7 & 4368.1 & 369.35 (738.7) & 2513.8 & 6884.2 & 9037.1 & No convergence \\ \hline
    Non-smooth dual & -279.69 & -279.69 & -279.69 & -279.69 & -259.28 & -258.27 & -258.7 & -260.94 \\ \hline
    Non-smooth primal & -279.67 & -279.69 & -279.66 & N/A & -261.04 & -258.36 & -258.86 & N/A \\ \hline
    Integer primal & -279.69 & -279.69 & -279.69 & -279.69 & -247.86 & -248.24 & -250.3 & -249.82 \\
    \hline
  \end{tabular}}
\caption{Results for synthetic problem.}
\label{synthTable}
\end{table*}

We next tested with matching points on the House dataset \cite{houseLINK}. $n=74$ points are marked in all the frames and the points in the first frame are matched with points in later frames (110 being the last frame). The unaries are set to zero. The camera rotates considerably around the object and we show results for three frames in table \ref{houseTable}.

\begin{table*}[t]
\resizebox{\textwidth}{!}{
  \centering
  \begin{tabular}{ | l | l | l | l | l | l | l | l | l | l | l | l | l | }
    \hline
     & \multicolumn{4}{c|}{Frame 70} & \multicolumn{4}{c|}{Frame 90} & \multicolumn{4}{c|}{Frame 110} \\ \hline
    Algorithm & TRN & SCD & FISTA & AD3 & TRN & SCD & FISTA & AD3 & TRN & SCD & FISTA & AD3 \\ \hline
    time (seconds) & 2374.5 & 3702.9 & 11964.5 & 1428.7 (2857.4) & 4731.6 & 4206.4 & 12561.2 & 2303.05 (4606.1) & 4451.8 & 10498.8 & 21171.1 & No convergence \\ \hline
    Non-smooth dual & -368.59 & -368.59 & -368.59 & -368.59 & -337.81 & -337.81 & -337.81 & -337.81 & -333.03 & -331.51 & -331.31 & -335.5 \\ \hline
    Non-smooth primal & -368.56 & -368.57 & -368.37 & N/A & -337.77 & -337.78 & -337.36 & N/A & -336.16 & -331.95 & -330.65 & N/A \\ \hline
    Integer primal & -368.59 & -368.59 & -368.59 & -368.59 & -337.81 & -337.81 & 337.81 & -337.81 & -315.93 & -317.69 & -317.78 & 314.16 \\
    \hline
  \end{tabular}}
\caption{Results for House dataset.}
\label{houseTable}
\end{table*}

\begin{figure}[t]
\centering
  \includegraphics[width=1.0\linewidth]{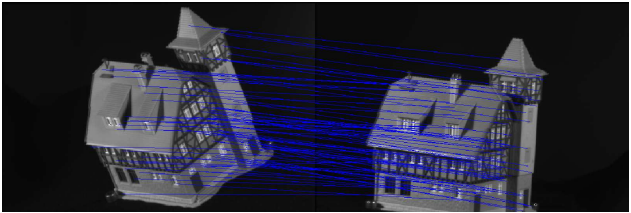}
\caption{Matching $1^{st}$ frame to $90^{th}$ frame.}
\label{fig:houseImg}
\end{figure}

The main reason for being able to tackle such problems is the efficient computation of log-sum-exp (section \ref{sparseSP}). {\sc trn-mrf}, {\sc fista} and {\sc scd}, all benefit from this. AD3 can also exploit sparse, pattern-based potentials, since, in its inner loop, max-product computations take place. The same steps shown for sum-product, can be used for max-product by replacing summing operation by max operation. Since, these modifications cannot be made to the AD3 version in openGM, we show within brackets the actual time taken by openGM's AD3. Since, our current implementation of sparse sum-product gives two times speed-up, the outside figure is half that value.

An important observation is that {\sc trn-mrf}, {\sc fista} and {\sc scd} have reliable convergence compared to {\sc ad3} (an {\sc admm} based approach). Among these three, {\sc trn-mrf} is very competitive and is the fastest in many cases. The quadratic convergence rate guarantee of {\sc trn-mrf} is evident because in many cases the stricter gradient based exit condition is reached at the same time or before the PD gap based exit condition. The gradient based exit condition is never reached before the PD gap condition by any of the first order methods. We do simple rounding of the primal variables to recover the labels. This rounding leads to energies that can be slightly better or worse. The crux of this paper is about getting closer to the global optimum of the non-smooth dual. Improved rounding schemes based on local search, will surely lead to better final labelling. AD3 has its own rounding scheme and outputs its results.

Next, we present results for stereo with curvature prior on $1 \times 3$ and $3 \times 1$ cliques. The clique energy is truncated, i.e., pattern-based. The unaries are based on absolute difference. We present results for Tsukuba, image size $144 \times 192$, 16 depth levels. This is a large problem and pattern-based sum-product was computed on clique chains. We compare quasi-Newton with {\sc fista} and {\sc ad3}. For large problems, the PD gap based criterion \cite{meshiChap14} never leads to convergence for both quasi-Newton and {\sc fista}. So, a simultaneous criterion on function value difference, variable difference and gradient has been used (adapted from \S 8.2.3.2 \cite{gillBOOK81}). AD3 showed poor convergence behavior for this large problem.

\begin{table}[!ht]
\centering
\resizebox{0.75\columnwidth}{!}{%
\begin{tabular}{ | l | l | l | l | }
    \hline
    Algorithm & Quasi-Newton & FISTA \\ \hline
    Iterations & 357 & 594 \\ \hline
    Non-smooth dual & 29105.9 & 29105.5  \\ \hline
    Integer primal & 29347 & 29282.5 \\
    \hline
\end{tabular}
}
\caption{Stereo estimation: Tsukuba.}
\label{tsuImg}
\end{table}
 
\begin{figure}[!ht]
\begin{center}
  \includegraphics[width=0.5\linewidth]{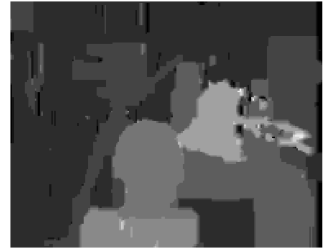}
\end{center}
   \caption{Tsukuba result for quasi-Newton.}
\label{fig:tsukuba_result}
\end{figure}

\section{Discussion}

We presented Newton-type methods that offer convergence guarantee and very good convergence rates, through appropriate choices made concerning their algorithmic components. Specifically, for problems in which sum-product computation is efficient, these Newton-type methods are very suitable. We showed promising results with higher-order {\sc mrf}s of medium and large sizes. We hope this work spurs further research on exploiting curvature information within optimization algorithms for {\sc map} inference.

\section{Acknowledgments}

Thanks to the reviewers for their helpful comments.

{\small
\bibliographystyle{ieee}
\bibliography{egbib}

\begin{thebibliography}{10}\itemsep=-1pt

\bibitem{csdLINK}
Complex step differentiation.
\newblock
  \url{http://blogs.mathworks.com/cleve/2013/10/14/complex-step-differentiation/}.
\newblock Accessed: 2016-10-30.

\bibitem{houseLINK}
House dataset.
\newblock \url{http://vasc.ri.cmu.edu//idb/html/motion/house/index.html}.
\newblock Accessed: 2016-10-30.

\bibitem{beckJIS09}
A.~Beck and M.~Teboulle.
\newblock A fast iterative shrinkage-thresholding algorithm for linear inverse
  problems.
\newblock {\em SIAM journal on imaging sciences}, 2(1):183--202, 2009.

\bibitem{bertsekasBook99}
D.~P. Bertsekas.
\newblock {\em Nonlinear programming}.
\newblock Athena scientific Belmont, 1999.

\bibitem{byrdSiamJO16}
R.~H. Byrd, S.~Hansen, J.~Nocedal, and Y.~Singer.
\newblock A stochastic quasi-newton method for large-scale optimization.
\newblock {\em SIAM Journal on Optimization}, 26(2):1008--1031, 2016.

\bibitem{byrdMP94}
R.~H. Byrd, J.~Nocedal, and R.~B. Schnabel.
\newblock Representations of quasi-newton matrices and their use in limited
  memory methods.
\newblock {\em Mathematical Programming}, 63(1-3):129--156, 1994.

\bibitem{connBook00}
A.~R. Conn, N.~I. Gould, and P.~L. Toint.
\newblock {\em Trust region methods}.
\newblock SIAM, 2000.

\bibitem{coughlanArxiv10}
J.~M. Coughlan and H.~Shen.
\newblock An embarrassingly simple speed-up of belief propagation with robust
  potentials.
\newblock {\em arXiv preprint arXiv:1010.0012}, 2010.

\bibitem{duchennePAMI11}
O.~Duchenne, F.~Bach, I.-S. Kweon, and J.~Ponce.
\newblock A tensor-based algorithm for high-order graph matching.
\newblock {\em IEEE transactions on pattern analysis and machine intelligence},
  33(12):2383--2395, 2011.

\bibitem{fixCVPR14}
A.~Fix, C.~Wang, and R.~Zabih.
\newblock A primal-dual algorithm for higher-order multilabel markov random
  fields.
\newblock In {\em Computer Vision and Pattern Recognition (CVPR)}, pages
  1138--1145. IEEE, 2014.

\bibitem{fountoulakisTR15}
K.~Fountoulakis and J.~Gondzio.
\newblock Performance of first-and second-order methods for big data
  optimization.
\newblock {\em arXiv preprint arXiv:1503.03520}, 2015.

\bibitem{gillBOOK81}
P.~E. Gill, W.~Murray, and M.~H. Wright.
\newblock Practical optimization.
\newblock 1981.

\bibitem{golubBook12}
G.~H. Golub and C.~F. Van~Loan.
\newblock {\em Matrix computations}, volume~3.
\newblock JHU Press, 2012.

\bibitem{hazanTransIT10}
T.~Hazan and A.~Shashua.
\newblock Norm-product belief propagation: Primal-dual message-passing for
  approximate inference.
\newblock {\em IEEE Transactions on Information Theory}, 56(12):6294--6316,
  2010.

\bibitem{ishikawaPAMI11}
H.~Ishikawa.
\newblock Transformation of general binary mrf minimization to the first-order
  case.
\newblock {\em IEEE Transactions on Pattern Analysis and Machine Intelligence},
  33(6):1234--1249, 2011.

\bibitem{jojicICML10}
V.~Jojic, S.~Gould, and D.~Koller.
\newblock Accelerated dual decomposition for map inference.
\newblock In {\em International Conference on Machine Learning (ICML)}, pages
  503--510, 2010.

\bibitem{surveyIJCV15}
J.~Kappes, B.~Andres, F.~Hamprecht, C.~Schnörr, S.~Nowozin, D.~Batra, S.~Kim,
  B.~Kausler, T.~Kröger, J.~Lellmann, N.~Komodakis, B.~Savchynskyy, and
  C.~Rother.
\newblock A comparative study of modern inference techniques for structured
  discrete energy minimization problems.
\newblock {\em International Journal of Computer Vision}, 115(2):155--184,
  2015.

\bibitem{kohliPAMI09}
P.~Kohli, M.~P. Kumar, and P.~H. Torr.
\newblock P$^3$ \& beyond: Move making algorithms for solving higher order
  functions.
\newblock {\em IEEE Transactions on Pattern Analysis and Machine Intelligence},
  31(9):1645--1656, 2009.

\bibitem{kollerBook09}
D.~Koller and N.~Friedman.
\newblock {\em Probabilistic graphical models: principles and techniques}.
\newblock MIT press, 2009.

\bibitem{kolmogorovPAMI15}
V.~Kolmogorov.
\newblock A new look at reweighted message passing.
\newblock {\em IEEE Transactions on Pattern Analysis and Machine Intelligence},
  37(5):919--930, 2015.

\bibitem{komodakisCVPR09}
N.~Komodakis and N.~Paragios.
\newblock Beyond pairwise energies: Efficient optimization for higher-order
  mrfs.
\newblock In {\em Computer Vision and Pattern Recognition (CVPR)}, pages
  2985--2992. IEEE, 2009.

\bibitem{komodakisPAMI11}
N.~Komodakis, N.~Paragios, and G.~Tziritas.
\newblock {MRF} energy minimization and beyond via dual decomposition.
\newblock {\em IEEE Transactions on Pattern Analysis and Machine Intelligence},
  33(3):531--552, 2011.

\bibitem{leeNIPS12}
J.~Lee, Y.~Sun, and M.~Saunders.
\newblock Proximal newton-type methods for convex optimization.
\newblock In {\em Advances in Neural Information Processing Systems (NIPS)},
  pages 836--844, 2012.

\bibitem{martensICML10}
J.~Martens.
\newblock Deep learning via hessian-free optimization.
\newblock In {\em International Conference on Machine Learning (ICML)}, pages
  735--742, 2010.

\bibitem{martinsJMLR15}
A.~F. Martins, M.~A. Figueiredo, P.~M. Aguiar, N.~A. Smith, and E.~P. Xing.
\newblock Ad3: Alternating directions dual decomposition for map inference in
  graphical models.
\newblock {\em Journal of Machine Learning Research}, 16:495--545, 2015.

\bibitem{meltzerUAI09}
T.~Meltzer, A.~Globerson, and Y.~Weiss.
\newblock Convergent message passing algorithms: a unifying view.
\newblock In {\em Uncertainty in Artificial Intelligence (UAI)}, pages
  393--401, 2009.

\bibitem{meshiMLKDD11}
O.~Meshi and A.~Globerson.
\newblock An alternating direction method for dual map lp relaxation.
\newblock In {\em Machine Learning and Knowledge Discovery in Databases}, pages
  470--483. Springer, 2011.

\bibitem{meshiChap14}
O.~Meshi, T.~Jaakkola, and A.~Globerson.
\newblock Smoothed coordinate descent for map inference.
\newblock {\em Advanced Structured Prediction}, pages 103--131, 2014.

\bibitem{miksikBMVC14}
O.~Miksik, V.~Vineet, P.~P{\'e}rez, P.~H. Torr, and F.~Cesson~S{\'e}vign{\'e}.
\newblock Distributed non-convex admm-inference in large-scale random fields.
\newblock In {\em British Machine Vision Conference (BMVC)}, 2014.

\bibitem{nocedalANL98}
J.~Nocedal and Y.-x. Yuan.
\newblock {\em Combining Trust Region and Line Search Techniques}, volume~14 of
  {\em Applied Optimization}, pages 153--175.
\newblock Springer, 1998.

\bibitem{rockafellarBook09}
R.~T. Rockafellar and R.~J.-B. Wets.
\newblock {\em Variational analysis}, volume 317.
\newblock Springer Science \& Business Media, 2009.

\bibitem{rotherCVPR09}
C.~Rother, P.~Kohli, W.~Feng, and J.~Jia.
\newblock Minimizing sparse higher order energy functions of discrete
  variables.
\newblock In {\em Computer Vision and Pattern Recognition, 2009. CVPR 2009.
  IEEE Conference on}, pages 1382--1389. IEEE, 2009.

\bibitem{savchynskyyCVPR11}
B.~Savchynskyy, S.~Schmidt, J.~Kappes, and C.~Schn{\"o}rr.
\newblock A study of nesterov's scheme for lagrangian decomposition and map
  labeling.
\newblock In {\em Computer Vision and Pattern Recognition (CVPR)}, pages
  1817--1823. IEEE, 2011.

\bibitem{savchynskyyUAI12}
B.~Savchynskyy, S.~Schmidt, J.~Kappes, and C.~Schn{\"o}rr.
\newblock Efficient mrf energy minimization via adaptive diminishing smoothing.
\newblock {\em Uncertainty in Artificial Intelligence (UAI)}, pages 746--755,
  2012.

\bibitem{schlickJCC87}
T.~Schlick and M.~Overton.
\newblock A powerful truncated newton method for potential energy minimization.
\newblock {\em Journal of Computational Chemistry}, 8(7):1025--1039, 1987.

\bibitem{schmidtPHD10}
M.~Schmidt.
\newblock {\em Graphical model structure learning using L1-regularization}.
\newblock PhD thesis, 2010.

\bibitem{shlezingerCYBER76}
M.~Shlezinger.
\newblock Syntactic analysis of two-dimensional visual signals in the presence
  of noise.
\newblock {\em Cybernetics and systems analysis}, 12(4):612--628, 1976.

\bibitem{sontagCHAP11}
D.~Sontag, A.~Globerson, and T.~Jaakkola.
\newblock Introduction to dual decomposition for inference.
\newblock {\em Optimization for Machine Learning}, 1:219--254, 2011.

\bibitem{sontagUAI08}
D.~Sontag, T.~Meltzer, A.~Globerson, T.~Jaakkola, and Y.~Weiss.
\newblock Tightening lp relaxations for map using message passing.
\newblock In {\em Uncertainty in Artificial Intelligence (UAI)}, pages
  503--510, 2008.

\bibitem{vineetIJCV14}
V.~Vineet, J.~Warrell, and P.~H. Torr.
\newblock Filter-based mean-field inference for random fields with higher-order
  terms and product label-spaces.
\newblock {\em International Journal of Computer Vision}, 110(3):290--307,
  2014.

\bibitem{wainwrightTRANSIT05}
M.~J. Wainwright, T.~S. Jaakkola, and A.~S. Willsky.
\newblock Map estimation via agreement on trees: message-passing and linear
  programming.
\newblock {\em IEEE Transactions on Information Theory}, 51(11):3697--3717,
  2005.

\bibitem{wangICML13}
H.~Wang and D.~Koller.
\newblock A fast and exact energy minimization algorithm for cycle mrfs.
\newblock In {\em Proceedings of the 30th International Conference on},
  volume~1, page~1, 2013.

\end{thebibliography}
}

\end{document}